\documentclass[runningheads]{llncs}

\usepackage{eccv}

\usepackage{eccvabbrv}

\usepackage{graphicx}
\usepackage{booktabs}

\usepackage[accsupp]{axessibility}  

\usepackage{amsmath,amssymb}
\usepackage{mathrsfs}
\usepackage{multirow}
\usepackage{colortbl}

\usepackage{hyperref}

\usepackage{orcidlink}

\begin{document}

\title{Neural Poisson Solver: A Universal and Continuous Framework for Natural Signal Blending} 

\titlerunning{Neural Poisson Solver}

\author{
Delong Wu\inst{1, \dag}\orcidlink{0009-0002-8416-4995} \ 
Hao Zhu\inst{1, \dag}\orcidlink{0000-0002-6756-9571} \ 
Qi Zhang\inst{2}\orcidlink{0000-0001-9611-6697} \ 
You Li\inst{3}\orcidlink{0000-0002-0152-1655} \\
Zhan Ma\inst{1, *}\orcidlink{0000-0003-3686-4057} \ 
Xun Cao\inst{1, *}\orcidlink{0000-0003-3094-4371}
}

\authorrunning{D.~Wu et al.}

\institute{
$^1$ Nanjing University \quad $^2$ Tencent AI Lab \\
$^3$ China Astronaut Research and Training Center
}

\maketitle

\let\thefootnote\relax\footnotetext{
    † \  These authors contributed equally to this work. \\
    * \  Corresponding authors: \email{\{caoxun, mazhan\}@nju.edu.cn}\\
    This work was supported by NSFC under Grant 62025108 and Tencent Rhino-Bird Joint Research Program RBFR2024009.
}

\begin{abstract}
Implicit Neural Representation (INR) has become a popular method for representing visual signals (\eg, 2D images and 3D scenes), demonstrating promising results in various downstream applications. Given its potential as a medium for visual signals, exploring the development of a neural blending method that utilizes INRs is a natural progression. Neural blending involves merging two INRs to create a new INR that encapsulates information from both original representations. A direct approach involves applying traditional image editing methods to the INR rendering process. However, this method often results in blending distortions, artifacts, and color shifts, primarily due to the discretization of the underlying pixel grid and the introduction of boundary conditions for solving variational problems. To tackle this issue, we introduce the Neural Poisson Solver, a plug-and-play and universally applicable framework across different signal dimensions for blending visual signals represented by INRs. Our Neural Poisson Solver offers a variational problem-solving approach based on the continuous Poisson equation, demonstrating exceptional performance across various domains. Specifically, we propose a gradient-guided neural solver to represent the solution process of the variational problem, refining the target signal to achieve natural blending results. We also develop a Poisson equation-based loss and optimization scheme to train our solver, ensuring it effectively blends the input INR scenes while preserving their inherent structure and semantic content. The lack of dependence on additional prior knowledge makes our method easily adaptable to various task categories, highlighting its versatility. Comprehensive experimental results validate the robustness of our approach across multiple dimensions and blending tasks. Project: \url{https://ep1phany05.github.io/NeuralPoissonSolver-website/}
\end{abstract}

\begin{figure}[tb]
\centering
\includegraphics[width=1.0\textwidth]{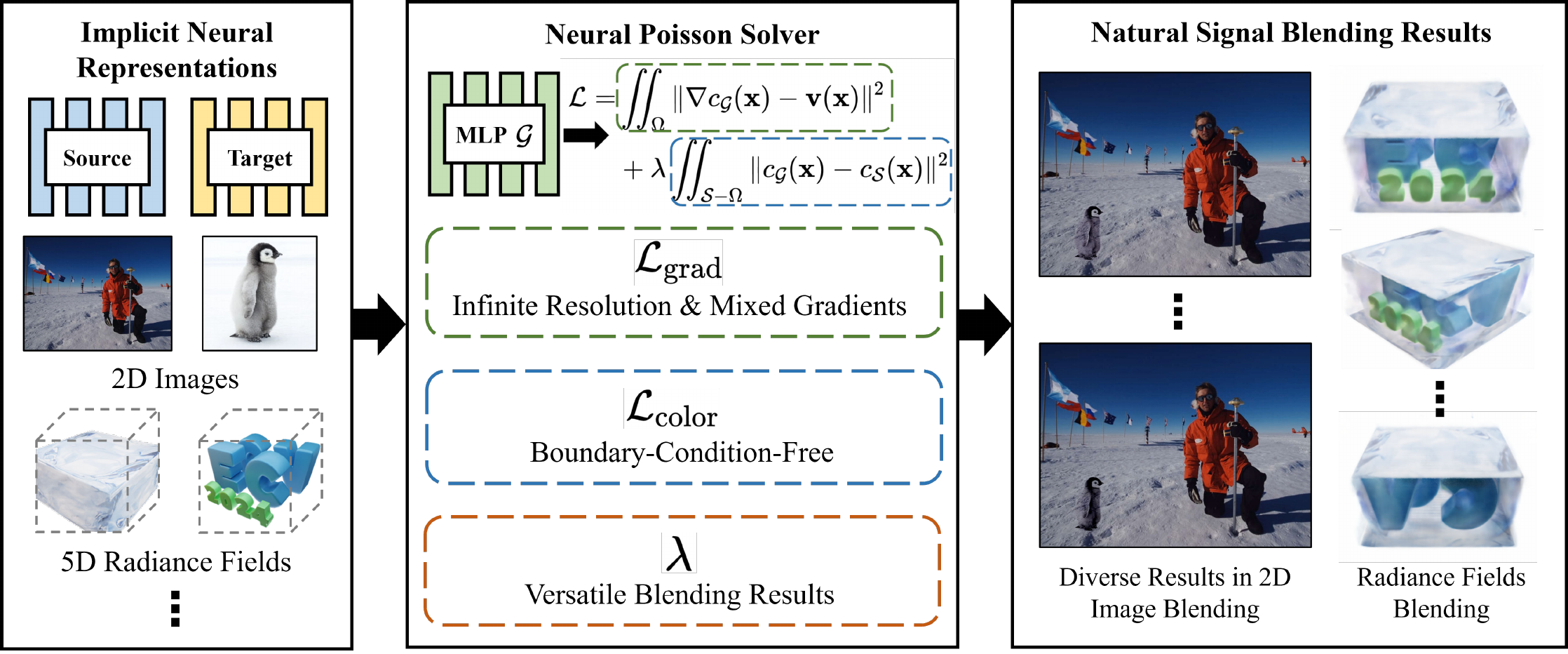}
\caption{\textbf{Overview.} Our method processes various dimensions of INRs effectively, without relying on prior knowledge. It utilizes a blending INR to carry the combined signal and introduces a Neural Poisson Solver for solving the Poisson equation during training and optimization of the blending outcome. In the Neural Poisson Solver, $\mathcal{L}_{\text{grad}}$ leverages the continuous nature of INR to provide gradients with infinite resolution, enhancing smoothness by integrating gradients from multiple directions, thereby minimizing the jaggedness of the blending result. Moreover, our approach obviates the need for traditional boundary conditions required in solving variational problems, employing $\mathcal{L}_{\text{color}}$ to broaden the receptive field of the blending area for a more seamless blending outcome. Finally, we introduce a hyperparameter $\lambda$ to fine-tune the balance between $\mathcal{L}_{\text{grad}}$ and $\mathcal{L}_{\text{color}}$, facilitating the achievement of varied and natural blending styles.}
\label{fig:poisson_1}
\end{figure}

\section{Introduction}
\label{sec:intro}

Implicit neural representation (INR) \cite{sitzmann2020implicit}, which characterizes the signal’s mapping function between the coordinates and attributes using neural networks, has been drawing increasing attention. Benefiting from the continuous function-based representation and convenient scalability to high dimensions, INR has been widely applied in various inverse optimization tasks across different dimensions \cite{mildenhall2021nerf, zhu2022dnf, liu2022recovery, zhou2023fourier, shen2021non, ruckert2022neat, xu2023nesvor, ma2023cardiacfield, raissi2020hidden}, inducing a new paradigm for signal processing. Consequently, developing novel tools for INR-based signal processing has become increasingly necessary.

We focus on the task of naturally blending given INRs, which is well-developed in classical matrix-based signal representation but rarely explored in the realm of INR. To achieve this goal, one straightforward method is to blend the signals in classical matrix-based representation by solving the Poisson partial differential equation (PDE) \cite{perez2023poisson}. However, due to the discrete signal representation, existing Poisson solvers (the Finite Difference approach \cite{perez2023poisson} and the Fourier approach \cite{morel2012fourier}) require proper boundary conditions, \ie, the Dirichlet boundary conditions, to produce a unique solution, resulting in distortion, artifacts, and color shifts in the blended results (see \cref{fig:boundary_condition}). Additionally, current solvers are specifically designed for processing 2D images. With the increase of signal dimensions and scales (\eg, 5D radiance fields \cite{mildenhall2021nerf}), the accuracy of classical solvers is significantly reduced \cite{karniadakis2021physics}, necessitating their re-implementation.

Such problems are closely related to the imperfections of the classical representation based on the underlying discrete pixel grid, where neighboring pixels are stored independently, and the continuous gradients are approximated by discrete differences. To address these issues, we propose the Neural Poisson Solver, built upon the continuous INR directly. Benefiting from INR's infinite resolution property, continuous gradients can be more accurately approximated, laying the groundwork for a superior solution to the partial differential equation. Furthermore, since INR can be easily expanded to signals with high dimensions (\ie, adding more input neurons), the Neural Poisson Solver's capability to process signals of various dimensions is ensured.

In the proposed Neural Poisson Solver, traditional computational math techniques are eschewed. Instead, the Poisson PDE is directly utilized as the loss function. Compared to classical solutions, the Neural Poisson Solver eliminates the need for Dirichlet boundary conditions, involving all points in the computation. As a result, blending outcomes are nearly unaffected by the mask's shape and complexity, achieving global blending details with enhanced robustness (see \cref{fig:boundary_condition}). Experiments on 2D image blending demonstrate that our solver reduces the Poisson PDE error to between $1/10$ and $1/1000$ of that of the traditional method \cite{perez2023poisson}, confirming the accuracy of our approach.Additionally, to demonstrate the solver's universal capability for processing signals of different dimensions, we applied it to the task of blending natural radiance fields, a topic seldom explored in literature, yielding blended radiance fields with barely noticeable effects.

Main contributions are summarized as follows:
\begin{itemize}
    \item We introduce the first editing tool for blending INRs, \ie, the Neural Poisson Solver, which operates without the need for prior knowledge or constraints. This offers unparalleled plug-and-play capabilities and facilitates easy adaptation to a wide range of signal types.
    \item The Neural Poisson Solver significantly reduces the error of the Poisson PDE to a range of $1/10$ to $1/1000$ compared to classical methods, marking a substantial improvement in accuracy.
    \item To the best of our knowledge, our work represents the inaugural effort to apply Poisson blending techniques within the domain of radiance fields, opening new avenues for research and application.
\end{itemize}

\section{Related Work}
\label{sec:related}

\noindent\textbf{2D Natural Signal Blending.} In 2D image processing, Poisson Image Editing \cite{perez2023poisson} emerges as a pivotal technique among gradient-domain methods. It notably enhances image blending by correcting color mismatches and ensuring smooth transitions, thereby significantly improving visual quality and appeal. Additionally, the Poisson equation serves as a versatile tool in addressing variational problems, finding effective applications across various fields such as computational photography, computer graphics, and machine vision \cite{fattal2023gradient, lewis2001lifting, elder1998image}, which underscores its broad utility and effectiveness.

While gradient-domain methods like Poisson Image Editing \cite{perez2023poisson} have markedly advanced 2D image processing, they are designed for specific tasks and depend on traditional approaches, encountering certain limitations. For instance, solving the Poisson equation typically involves the finite difference method, as initially proposed by Pérez \etal \cite{perez2023poisson}, or the Fourier method \cite{morel2012fourier}, leveraging Fourier transform properties. However, both methods rely on image discretization, which can irreversibly alter image blending outcomes.

Furthermore, these techniques assume certain conditions about the input data, limiting their applicability in complex situations. They typically ignore depth and occlusion, crucial in 3D scene understanding and reconstruction, leading to less than optimal outcomes in tasks requiring spatial relationship comprehension or dealing with occlusions, where discerning visible and obscured scene parts is essential.

\noindent\textbf{Implicit Neural Representations (INRs)} \cite{sitzmann2020implicit} have revolutionized computer vision and graphics by providing a continuous mapping between coordinates and attributes through neural networks. This approach allows for capturing intricate patterns in both 2D and 3D, offering a seamless representation of scenes. Unlike traditional high-resolution images, 3D meshes, or point clouds, INRs are resolution-independent and more compact, enhancing their versatility across various applications. These include media representation and compression \cite{gao2022objectfolder, strumpler2022implicit}, reconstruction and rendering in vision and graphics \cite{mildenhall2021nerf, tewari2022advances, zhu2023pyramid}, advancements in microscopy through holography and tomography \cite{zhu2022dnf, liu2022recovery}, materials science via meta-surface design \cite{chen2020physics}, computational mathematics for solving differential equations \cite{raissi2019physics, karniadakis2021physics}, and hydrodynamics for fluid simulation \cite{raissi2020hidden}. This development signifies a major shift in signal processing, indicating a new era with wide-ranging implications.

Despite image blending being a well-researched topic in computational photography, few works have explored blending tasks from the INRs perspective. This is primarily due to the challenge of transferring appearance information between source and target scenes as feature information in Neural Radiance Fields is encoded within black-box network parameters.

\noindent\textbf{Challenges in 3D Signal Blending.} Regarding natural editing on 3D objects, several methods have been proposed for texture stitching and blending \cite{rocchini1999multiple,dessein2014seamless}. However, these texture-based methods are not suitable for NeRF models. Recent studies have started to explore editing NeRF scenes \cite{wang2022clip,liu2021editing}. The initial efforts in this area introduced a version of NeRF that depends on implicit shape and appearance encodings, enabling separate editing of the shape and color of 3D objects. Nonetheless, these approaches encounter challenges with accurately identifying specific areas for editing and providing flexible editing options.

Recent studies have explored various aspects of editing in NeRF models, including geometric editing \cite{kobayashi2022decomposing, wu2022palettenerf}, global style changes \cite{chen2022upst, chiang2022stylizing, fan2022unified, huang2022stylizednerf}, recoloring \cite{gong2023recolornerf}, and scene separation for localized edits \cite{yuan2022nerf, benaim2024volumetric}. Despite these advancements, there's room for improvement in accurately identifying specific editing targets and in the adaptability of these methods. Additionally, the operations demonstrated by these studies are relatively basic, focusing on tasks like object removal and simple editing techniques.

In conclusion, while there has been significant research in blending natural signals, existing approaches encounter challenges in handling blending tasks across different dimensions and signal types, particularly when applying these methods to NeRF models. Traditional techniques often lack the necessary design specificity and struggle to offer a versatile framework. They also tend to fall short in addressing occlusions and ensuring view consistency in 3D environments. Thus, the development of a more comprehensive and adaptable framework, as proposed in this paper, is crucial.

\section{Preliminary}
\label{sec:pre}
\subsection{Poisson Editing Theory}
\label{sec:pre_1}
Poisson Image Editing (PIE), first proposed by Pérez \etal \cite{perez2023poisson}, introduced a novel method for blending images using variational techniques and the Poisson equation. This section outlines the mechanism of traditional Poisson-based image blending methods, which serve as an inspiration for our subsequent theory. In the PIE method, blending the target image $\mathcal{T}$ into the source image $\mathcal{S}$ can be formalized as a variational problem:
\begin{equation}
\label{eq:poisson_1}
    \min _{f \in \mathscr{C}^{2}(\mathcal{S})} \iint_{\Omega}\|\nabla f-\mathbf{v}\|^{2} d x,
    \text{ with } \left.f\right|_{\partial \Omega}=\left.f^{*}\right|_{\partial \Omega}, 
\end{equation}
where $\mathcal{S}$ is a closed subset of $\mathbb{S}^{2}$ representing the image domain, $\mathscr{C}^{2}(\mathcal{S})$ denotes twice differentiable real functions over $\mathcal{S}$'s interior, $\Omega \subset \mathcal{S}$, $f^{*}$ is the background image, and $\mathbf{v}$ is a differentiable gradient field from the selected region. Typically, better blending results can be obtained when $\mathbf{v} = \arg\max_{f^{*},g} \{ |\nabla f^{*}(\mathbf{x})|, |\nabla g(\mathbf{x})| \}$ or $\mathbf{v}= \mu \nabla f^{*}(\mathbf{x}) + \varphi \nabla g(\mathbf{x})$.

In Poisson equation-based image blending task, $\mathbf{v}$ refers to the combined gradient of the target and source images. Thus, the goal of solving equation (\ref{eq:poisson_1}) is to: (1) align $\nabla f$ with $\mathbf{v}$ within $\Omega$, and (2) align $f$'s boundary value with source image $f^{*}$ at the edge of $\Omega$ (\ie, $\partial \Omega$). To solve this variational equation, a boundary condition is added. Given that it satisfies the Euler-Lagrange equation, Pérez \etal proposed solving under Dirichlet boundary conditions and converted formula (\ref{eq:poisson_1}) into the Poisson equation:
\begin{equation}
\label{eq:poisson_2}
    \Delta f(x)=\operatorname{div}(\mathbf{v}(x)) 
    \text { for all } x \in \Omega, 
    \text { and }\left.f\right|_{\partial \Omega}=\left.f^{*}\right|_{\partial \Omega}. 
\end{equation}

\subsection{Neural Radiance Fields}
Neural Radiance Fields (NeRF)\cite{mildenhall2021nerf} generate images by sampling 5D coordinates, which include spatial location $(x, y, z)$ and viewing direction $(\theta, \phi)$. These samples are then mapped to color $(r, g, b)$ and volume density $\sigma$ along camera rays. This function is characterized using coordinate-based Multilayer Perceptron (MLP) networks \cite{sitzmann2020implicit}. Subsequently, volumetric rendering techniques are utilized to alpha composite the values at each point, thus producing the final images.Let's consider a pixel $r(t) = o + td$, where $o$ is the camera origin and $d$ is the ray direction. The predicted color for this pixel can be defined as:
\begin{equation}
\label{eq:nerf_1}
    C(r)=\int_{t_{n}}^{t_{f}} \exp \left(-\int_{t_{n}}^{t} \sigma(s) ds\right) \sigma(t) c(t, d) dt, 
\end{equation}
where $t_{n}$ and $t_{f}$ define near and far bounds, $\sigma(\cdot)$ and $c(\cdot, \cdot)$ denote densities and color predictions from the network respectively. Due to computational constraints, the continuous integral is numerically approximated using quadrature. Finally, NeRF optimizes the radiance field by minimizing the mean squared error between rendered color and ground truth color expressed.

\begin{figure}[tb]
\centering
\includegraphics[width=0.95\textwidth]{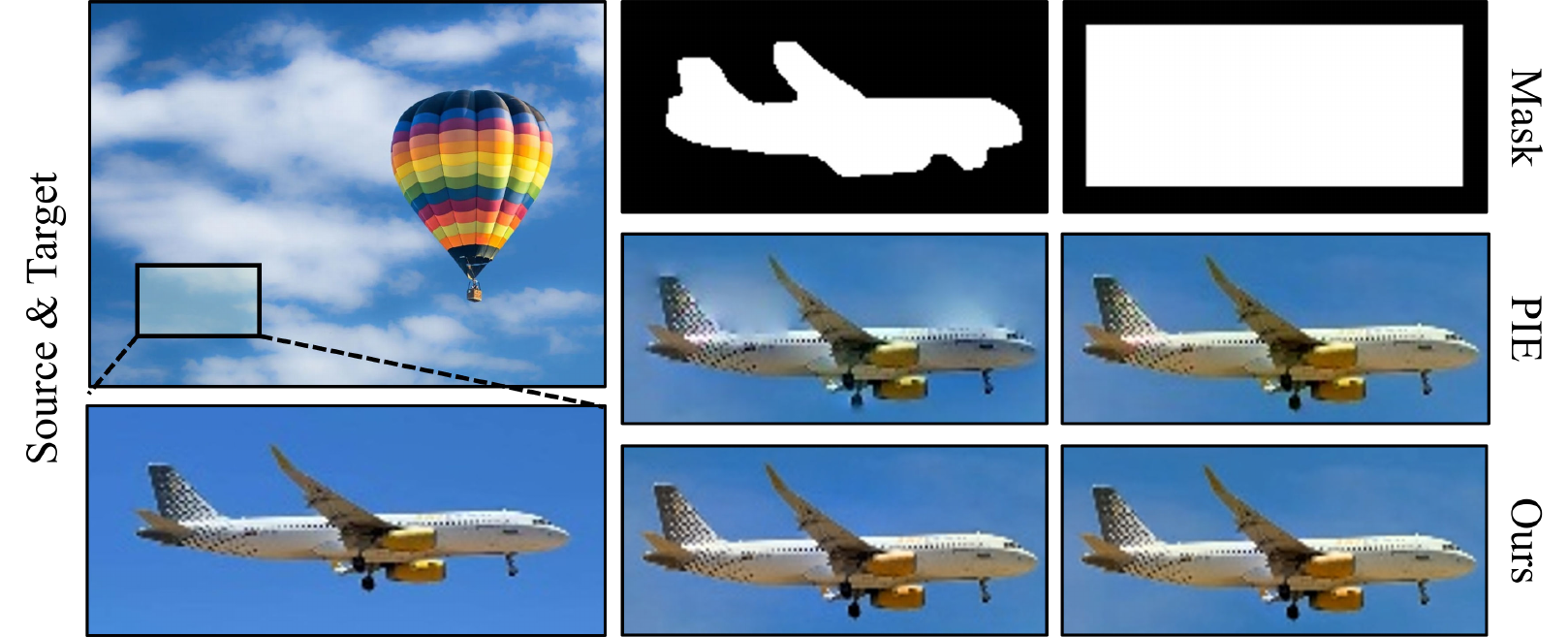}
\caption{In PIE \cite{perez2023poisson}, the selection of mask shapes significantly influences the final blending outcomes, particularly regarding texture and color details. Masks that are too closely positioned can result in color bleeding artifacts within the synthesized image areas. In contrast, our method, which does not rely on boundary conditions, is not impacted by the proximity of mask shapes.}
\label{fig:boundary_condition}
\end{figure}

\section{Neural Poisson Solver}
\label{sec:method_2d}
\subsection{Representing Signals as Continuous INR}
INRs \cite{sitzmann2020implicit} encapsulate signal attributes by interpreting the signal as a function of its corresponding coordinates. INR establishes a relationship between coordinates and their respective signal values, making it suitable for the continuous and memory-efficient modeling of a wide range of signals, such as 1D audio \cite{gao2022objectfolder}, 2D images \cite{tancik2020fourier}, 3D shapes \cite{park2019deepsdf}, 4D light fields \cite{sitzmann2021light}, and 5D radiance fields \cite{mildenhall2021nerf}. Consider $\mathbf{x} \in \mathbb{R}^{M}$ as the input coordinates and $\mathbf{f} \in \mathbb{R}^{N}$ as the corresponding output feature values. In the context of INR, we define:
\begin{equation}
\label{eq:inr_1}
    \mathbf{f} 
    =\varphi(\mathbf{x}) 
    =\mathbf{w}_{n}(\phi_{n-1} \circ \phi_{n-2} \circ ... \circ \phi_{0})(\mathbf{x})
    +\mathbf{b}_{n},
\end{equation}
where each $\phi_{i}(\mathbf{x}) = \mathcal{F}(\mathbf{w}_{i} \cdot \mathbf{x} + \mathbf{b}_{i})$ represents the mapping function of the $i^{th}$ layer in the neural network. Here, $\mathcal{F}$ denotes the activation function, and $\mathbf{w}_{i}$ and $\mathbf{b}_{i}$ are the weight matrix and bias term of the $i^{th}$ layer, respectively.

Traditional signal representation approaches typically use a discrete format. As a result, both the signal itself and any finite element solution strategies based on the signal can suffer from potential distortions due to the discretization steps involved. In contrast, INR parameterizes the signal as a continuous function. This method maps the signal domain to attribute values at corresponding coordinates, offering improved resilience to spatial resolution constraints.

In our work, we represent discrete signals—specifically, source and target images—using two independent INRs, referred to as $\mathcal{S}$ and $\mathcal{T}$ for consistency with previous descriptions. Although INR is capable of representing signals in a continuous manner, its capacity is limited by the architecture of the underlying network model. To represent and blend more complex signals that contain higher-frequency details, we utilize the Disorder-Invariant Implicit Neural Representation (DINER) \cite{zhu2023disorder}. DINER initially maps input coordinates to a new index using a hash table. These mapped geometric coordinates are then processed by the conventional INR backbone network. Through adaptive indexing with the hash table, DINER ensures that the same signal mapping can encapsulate more low-frequency components, regardless of the arrangement of signal elements.

\begin{figure}[tb]
\centering
\includegraphics[width=1.0\textwidth]{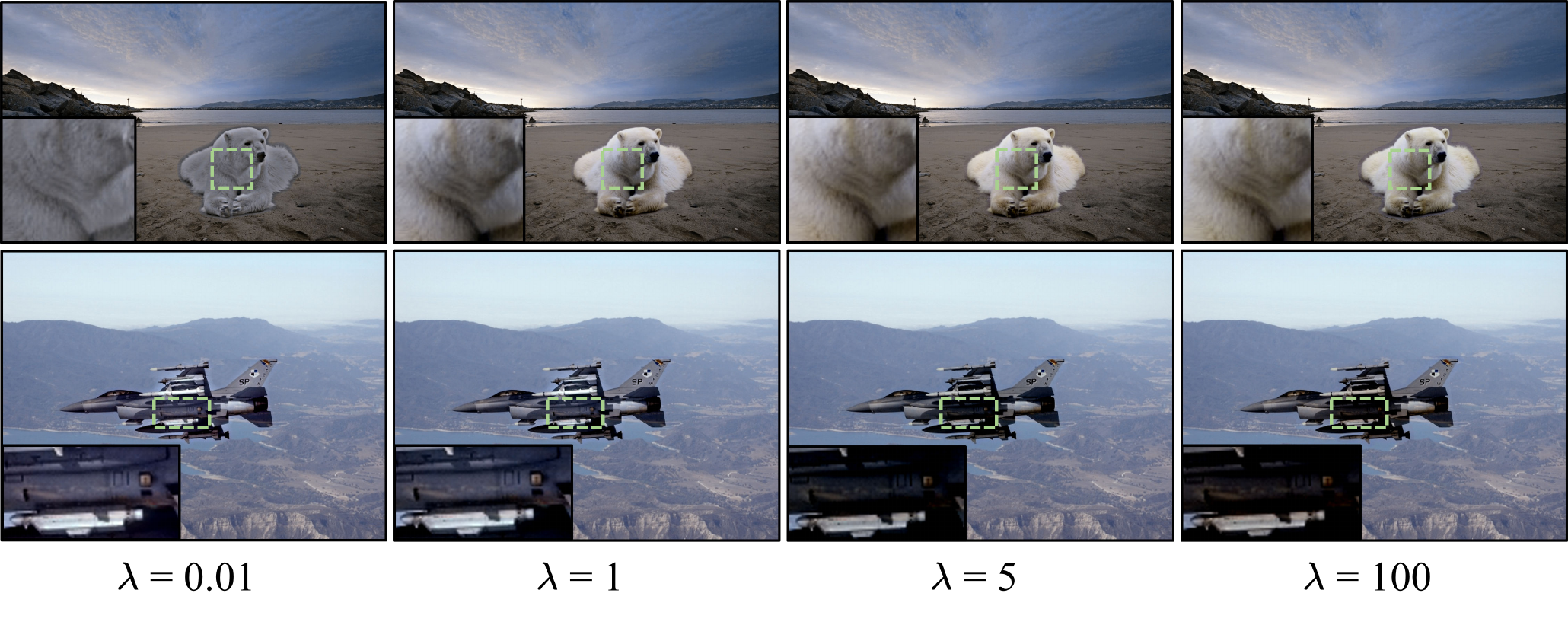}
\caption{\textbf{Neural Blending Operator.} When $\lambda<1$, the blending operation accentuates the gradient within the blending region $\Omega$, thereby more effectively preserving the intricate details of the scene. However, this approach may lead to a noticeable color discrepancy between the blending edge $\partial \Omega$ and the background INR $\mathcal{S}$. On the other hand, as $\lambda$ increases, the focus shifts towards ensuring a smooth transition at $\partial \Omega$. While this method facilitates a seamless blend, it might slightly diminish the precision of the scene's detailed features.}
\label{fig:2d_result_1}
\end{figure}

\subsection{Boundary-Condition-Free Poisson Solver}
For 2D image blending, the PIE method \cite{perez2023poisson} requires specific boundary conditions to solve the variational problem \cref{eq:poisson_1} for a unique solution. Despite its significant contributions to image editing, this method faces some challenges. It converts the variational problem \cref{eq:poisson_1} into the Poisson equation \cref{eq:poisson_2} using the Euler-Lagrange equation, depending on Dirichlet boundary conditions for a unique solution. The choice of boundary conditions critically affects the blending results. A precise mask around the foreground improves object-background distinction and pixel value calculations in the blending zone. However, an overly close mask to the target object may cause color bleeding artifacts in the mismatched blending area. As shown in \cref{fig:boundary_condition}, mask variations impact the results of the PIE method \cite{perez2023poisson}, whereas our method is unaffected. The limitation of Dirichlet boundary conditions to the edges of the blending area, neglecting the global background, can lead to color inaccuracies \cref{fig:ablation}. Benefiting from the continuity and differentiability of INR, our solution to the Poisson equation is not limited by boundary conditions. Thus, we extend the blending task to the entire background $(\mathcal{S} - \Omega)$, aiming for a more comprehensive blending effect and reducing color blending errors common in traditional methods. We introduce a blank INR $\mathcal{G}$ that conforms to the specifications of $\mathcal{S}$ and $\mathcal{T}$.

This endeavor seeks to naturally integrate a region of interest (ROI) $\Omega$ from the target signal $\mathcal{T}$ into a specific source signal $\mathcal{S}$ location, focusing on two main aspects. The first, $L_{\text{grad}}$, aligns function $f$'s gradient with the guiding vector field $\mathbf{v}$ within $\Omega$, preserving key features of $\mathcal{S}$ and $\mathcal{T}$. The second, $L_{\text{color}}$, ensures $\partial f$ and $\partial f^{*}$ consistency on $\partial \Omega$, keeping pixel values inside $\Omega$ similar to the background $\mathcal{S}$ and preventing harsh transitions:
\begin{subequations}
\begin{align}
    \mathcal{L}_{\text{grad}} 
    &=\iint_{\Omega}\|\nabla c_{\mathcal{G}}(\mathbf{x})-\mathbf{v}(\mathbf{x})\|^{2},  \label{eq:loss_g} \\
    \mathcal{L}_{\text{color}} 
    &=\iint_{\mathcal{S}-\Omega}\|c_{\mathcal{G}}(\mathbf{x})-c_{\mathcal{S}}(\mathbf{x})\|^{2}.  \label{eq:loss_b}
\end{align}
\end{subequations}

We introduce a hyperparameter $\lambda$ to balance the weights of $\mathcal{L}_{\text{grad}}$ and $\mathcal{L}_{\text{color}}$, facilitating varied blending effects. The optimization of $\mathcal{G}$ via $\mathcal{L} = \mathcal{L}_{\text{grad}} + \lambda \mathcal{L}_{\text{color}}$ over multiple iterations yields superior blending results. A higher $\lambda$ value prompts the Neural Poisson Solver to prioritize the smooth transition between the blending edge, $\partial \Omega$, and the background INR, $\mathcal{S}$. As a result, feature clarity within the blending area, $\Omega$, aligning with the target INR, $\mathcal{T}$, diminishes, as intuitively shown in \cref{fig:2d_result_1}.

\section{Experiment}
\label{sec:exp}
In \cref{sec:compare}, we initially present both qualitative and quantitative comparisons of our proposed method against traditional approaches that utilize discretized solutions to the Poisson equation, specifically for 2D image blending. Following this, in \cref{sec:result}, we showcase the effectiveness of our method in blending 5D Radiance Fields. In \cref{sec:ablation}, we highlight the impact of our Boundary-Condition-Free method on improving fidelity and visual quality. Finally, in \cref{sec:comparison}, we qualitatively compare our method with several state-of-the-art techniques in image editing and blending.

\begin{table}[tb]
    \centering
    \scriptsize
    \caption{\textbf{2D image blending results.} Quantitative comparison with the discretized PIE method \cite{perez2023poisson} and ours. $\mathcal{L}/N$ represents calculating the average error on each pixel.}
    \label{tab:quantitative}
    \begin{tabular*}{\textwidth}{@{\extracolsep{\fill}}ccccccc} 
        \toprule
        \multicolumn{2}{c}{Scene} & \multicolumn{1}{c}{\multirow{2}{*}{Solver}} & \multicolumn{2}{c}{$\arg\max_{f^{*},g} \{ |\nabla f^{*}(\mathbf{x})|, |\nabla g(\mathbf{x})|$} & \multicolumn{2}{c}{$\mu \nabla f^{*}(\mathbf{x}) + \varphi \nabla g(\mathbf{x})$} \\ 
        \cmidrule{4-7}
         Source & Target &  & $(\mathcal{L}/N)_{\text{grad}} \downarrow$ & $(\mathcal{L}/N)_{\text{color}} \downarrow$ & $(\mathcal{L}/N)_{\text{grad}} \downarrow$ & $(\mathcal{L}/N)_{\text{color}} \downarrow$ \\
        \midrule
         \multirow{2}{*}{Snowfield}
            &\multirow{2}{*}{Penguin} & PIE & 0.5071 & 1.4942 & 0.4429 & 1.2461 \\
            & & Ours & \textbf{0.0514} & \textbf{0.0011} & \textbf{0.0021} & \textbf{0.0015} \\ 
        \midrule
         \multirow{2}{*}{Sky}
            &\multirow{2}{*}{Aircraft} & PIE & 0.7635 & 2.2599 & 0.7621 & 1.4981 \\
            & & Ours & \textbf{0.0026} & \textbf{0.0024} & \textbf{0.0028} & \textbf{0.0024} \\ 
        \midrule
         \multirow{2}{*}{Board}
            &\multirow{2}{*}{Slogan} & PIE & 32.495 & 6.6799 & 31.2751 & 7.0393 \\
            & & Ours & \textbf{0.1744} & \textbf{0.0026} & \textbf{0.2189} & \textbf{0.0027} \\ 
        \midrule
         \multirow{2}{*}{Sky}
            &\multirow{2}{*}{Fighter} & PIE & 3.7561 & 5.5416 & 3.6714 & 1.6783 \\
            & & Ours & \textbf{0.0123} & \textbf{0.0030} & \textbf{0.0164} & \textbf{0.0030} \\ 
        \midrule
         \multirow{2}{*}{Brick}
            &\multirow{2}{*}{Letter} & PIE & 7.3058 & 8.3664 & 5.5999 & 8.2730 \\
            & & Ours & \textbf{0.0316} & \textbf{0.0391} & \textbf{0.0415} & \textbf{0.0390} \\ 
        \bottomrule  
    \end{tabular*}
\end{table}

\begin{figure}[tb]
\centering
\includegraphics[width=1.0\textwidth]{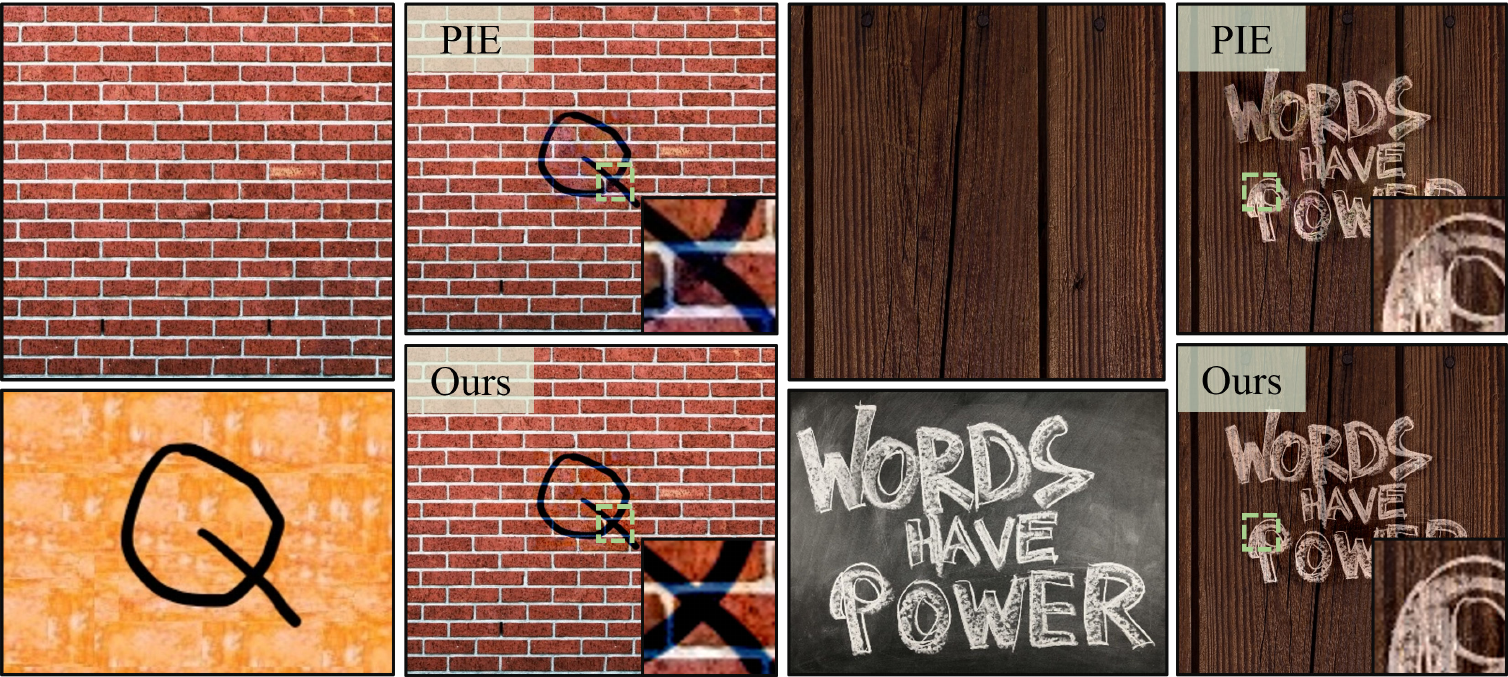}
\caption{Displaying the blending results of the PIE method \cite{perez2023poisson}  and our approach across different 2D scenes. The first and third columns show the source and target scenes for two tasks, respectively. The second and fourth columns respectively showcase the blending outcomes and related details.}
\label{fig:2d_result_2}
\end{figure}

\begin{figure}[tb]
\centering
\includegraphics[width=1.0\textwidth]{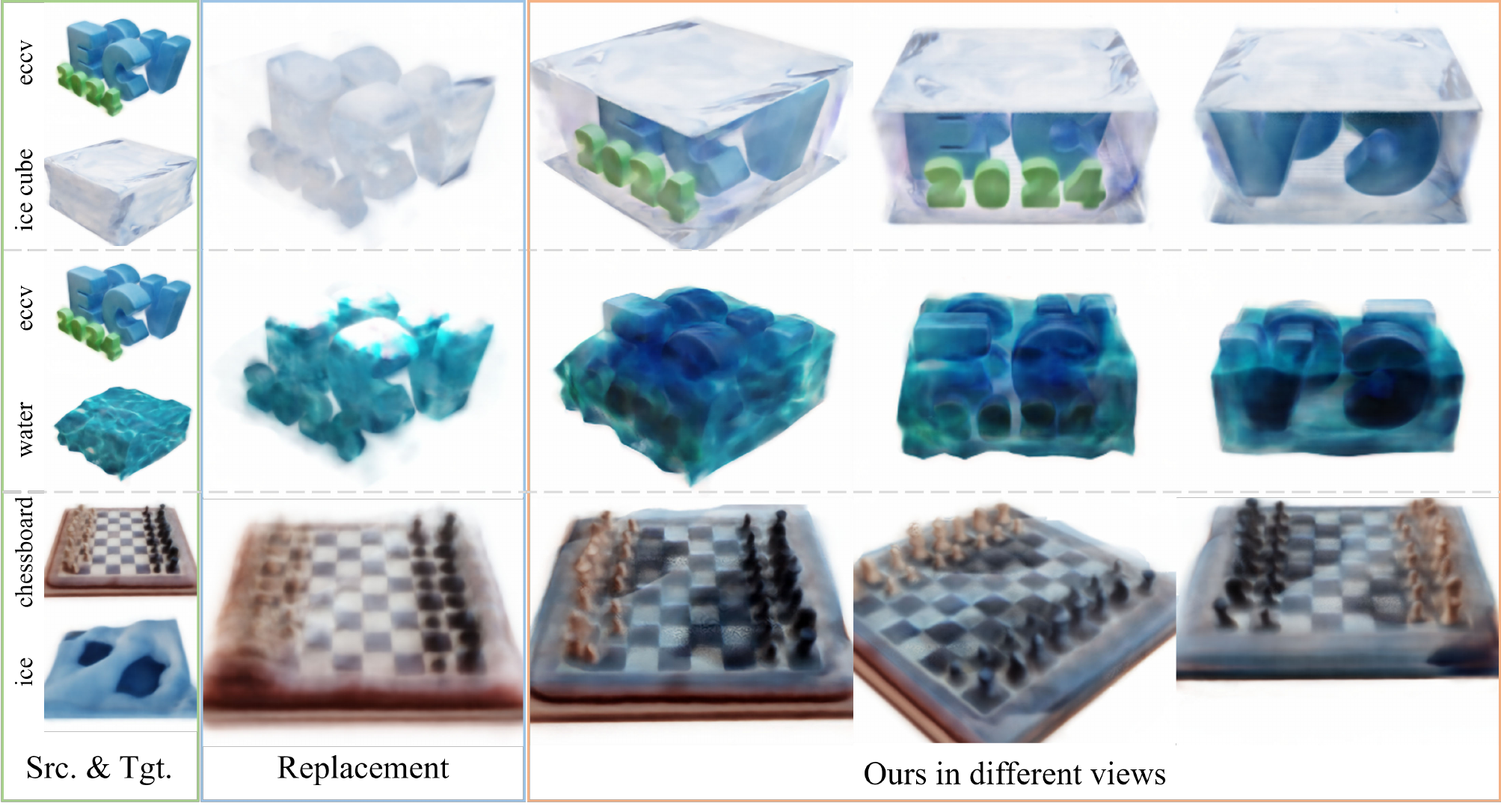}
\caption{Our method in Radiance Fields blending. The first column shows the original scenes. The second column employs a common replacement blending approach, directly substituting the region of interest $\Omega$ in $\mathcal{S}_{\Theta}$ with $c_{\mathcal{G}}(\mathbf{x})$, rendered based on $\alpha_{\mathcal{G}}(\mathbf{x})$. Columns three to five demonstrate the blending results achieved with our method, showcasing the naturalness and consistency of the blend from different perspectives.}
\label{fig:3d_result}
\end{figure}

\subsection{2D Image Blending Task}
\label{sec:compare}

\noindent\textbf{Tasks.} The experimental design aimed to assess extensive region blending, instances where the gradient of the source image closely aligns with the edge gradient of the target, and scenarios involving minimal to maximal gradient variations in the blending region of the source scene. 

\noindent\textbf{Configurations.} For the 2D image blending task, we adopted the same framework as outlined in DINER\cite{zhu2023disorder}. We introduce source INR $\mathcal{S}$ and target INR $\mathcal{T}$ based on different scene dimensions, and initialize an INR $\mathcal{G}$ of the same specifications to store the blending results. We utilized the two most commonly used operators from PIE \cite{perez2023poisson} for the image blending task as guiding vector fields. These operators are defined as $\mathbf{v} = \arg\max_{f^{*},g} \{ |\nabla f^{*}(\mathbf{x})|, |\nabla g(\mathbf{x})| \}$ and $\mathbf{v}= \mu \nabla f^{*}(\mathbf{x}) + \varphi \nabla g(\mathbf{x})$, with both $\mu$ and $\varphi$ set to 1 in our experiments. To ensure the fairness of our experiments, we used a moderately sized, well-positioned rectangular mask to define the blending area, avoiding unnecessary errors in PIE \cite{perez2023poisson}.

Our framework training was conducted on a single NVIDIA RTX 3090 GPU, using the Adam optimizer\cite{kingma2017adam} with a cosine annealing learning rate reduction strategy. The entire process took approximately $50$ to $100$ seconds. Our experiments were carried out on a series of representative 2D scenes, with both qualitative and quantitative metric evaluations performed. Our method shares the same goal with traditional methods that employ discretized solutions to Poisson's equation—to optimize the solution of the variational problem specified in \cref{eq:poisson_1}, primarily composed of \cref{eq:loss_g} and \cref{eq:loss_b}. For a fair comparison, we adjusted these equations to calculate average errors instead of total errors. These modified equations served as quantitative indicators for evaluating the efficacy of blending—\cref{eq:loss_g} indicates gradient feature retention, and \cref{eq:loss_b} reflects the smoothness of color transitions.

\noindent\textbf{Results.} \cref{tab:quantitative} shows a comparison between our Neural Poisson Solver and the PIE method \cite{perez2023poisson}. Our model outperforms the traditional PIE method significantly across all metrics, with improvements ranging from $10\times$ to $1000\times$. This indicates our model's superior ability in blending 2D images. Our approach also ensures minimal impact on the surrounding scene while maintaining the essential gradient features in the blend area.

\cref{fig:2d_result_2} showcases a comparison of blending results between our proposed Neural Poisson Solver and the traditional PIE method \cite{perez2023poisson}, with a focus on the detailed features of the blend area. The images reveal that the PIE method \cite{perez2023poisson} can lead to significant color shifts or errors in the blended objects, or introduce unnecessary redundant information into the background. Our approach solves the Poisson equation more accurately than the PIE method and captures the global background information of the image more effectively by expanding the receptive field.

\begin{figure}[tb]
\centering
\includegraphics[width=1.0\textwidth]{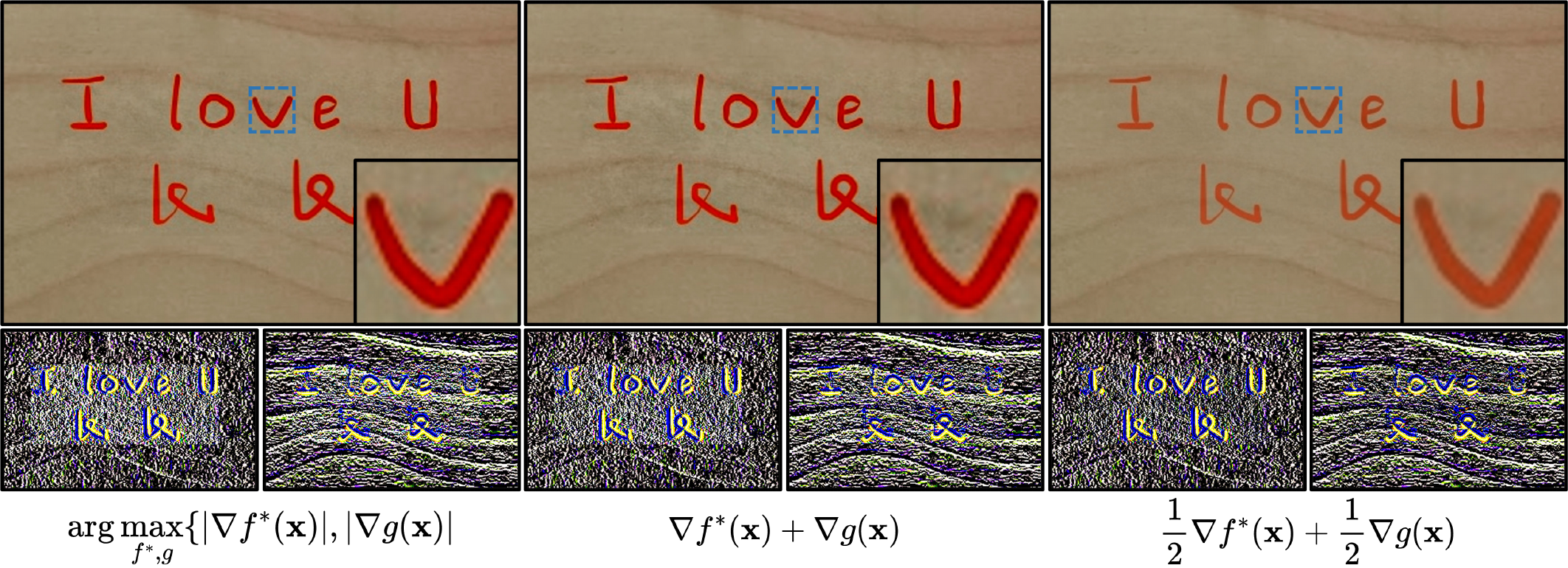}
\caption{Illustrate the impact of different guiding vector fields $\mathbf{v}$ on the mixed gradients in the $x$ (bottom left) and $y$ (bottom right) directions, as well as the final blending outcome (top): (1) feature blending, (2) feature overlay ($\mu = \varphi = 1$), (3) feature smoothing between source and target scenes ($\mu = \varphi = \frac{1}{2}$).}
\label{fig:different_v}
\end{figure}

\subsection{5D Radiance Fields Blending Task}
\label{sec:result}

\noindent\textbf{Tasks.} To our knowledge, no work has yet achieved the task of naturally and seamlessly blending two 5D Radiance Fields, with most efforts based on NeRF focusing on basic editing, generation, and stitching. We have applied our Neural Poisson Solver to NeRF, making preliminary attempts at heterogeneous and large-scale blending of 5D Radiance Fields.

\noindent\textbf{Methods.} NeRF\cite{mildenhall2021nerf} models a scene as a 5D function that maps 3D coordinates $\mathbf{x} = (x, y, z)$ and 2D viewing directions $\mathbf{d} = (\theta, \phi)$ to color $\mathbf{c} = (r, g, b)$ and density $\sigma$. This function is simulated by an MLP, $F_{\Theta}:(\mathbf{x}, \mathbf{d}) \rightarrow(\mathbf{c}, \sigma)$. In NeRF, density $\sigma$ depends only on $\mathbf{x}$, facilitating the use of the Neural Poisson Solver for blending in 5D Radiance Fields.

To blend source and target 5D Radiance Fields, $\mathcal{S}_{\Theta}$ and $\mathcal{T}_{\Theta}$, we select a 3D Region of Interest (ROI) $\Omega$ within $\mathcal{T}_{\Theta}$ and a central coordinate $p$ in $\mathcal{S}_{\Theta}$. The objective is to seamlessly integrate $\Omega$ from $\mathcal{T}_{\Theta}$ into $\mathcal{S}_{\Theta}$ around $p$. We start with a copy of $\mathcal{S}_{\Theta}$, named $\mathcal{G}_{\Theta}$, and make adjustments for blending. The modifications in $\mathcal{G}_{\Theta}$ focus on areas defined by $\Omega$ and $p$, while the rest is preserved through ray rendering within $\Omega$. For smooth blending, we use uniform rays across $\mathcal{S}_{\Theta}$, $\mathcal{T}_{\Theta}$, and $\mathcal{G}_{\Theta}$. Outside $\Omega$, we rely on outputs from $\mathcal{S}_{\Theta}$; inside, we blend results from the three fields, guided by $\mathbf{v}$ in the Neural Poisson Solver. 
Inspired by Blended-NeRF \cite{gordon2023blended}, optimization focuses on 3D points within $\Omega$ to save memory, rendering along rays through $\Omega$ and setting external $\sigma$ to 0 as shown in \cref{eq:optim_1}:
\begin{equation}
\label{eq:optim_1}
    C(\boldsymbol{r}) 
    =I(\exists x_{i} \in \boldsymbol{r}, x_{i} \in \Omega) \sum_{x_{i} \in \Omega} T_{i}\left(1-\exp \left(-\sigma_{i} \delta_{i}\right)\right) c_{i}. 
\end{equation}

Post-training, scenes within and outside $\Omega$ blend using the same rays, with $\mathcal{T}_{\Theta}$ and $\mathcal{G}_{\Theta}$ rendering inside points, and $\mathcal{S}_{\Theta}$ outside, guided by $\mathbf{v}$. For $\mathbf{v} = \mu \nabla c_{\mathcal{S}}(\mathbf{x}) + \varphi \nabla c_{\mathcal{T}}(\mathbf{x})$, it aligns with \cref{eq:alpha} and \cref{eq:color}:
\begin{subequations}
\begin{align}
    \sigma(\mathbf{x})
    &=\mathcal{F}(\mu\sigma_{\mathcal{G}}(\mathbf{x})
    +\varphi\sigma_{\mathcal{T}}(\mathbf{x})), \label{eq:alpha} \\
    c(\mathbf{x})
    &=\frac{\mu c_{\mathcal{G}}(\mathbf{x}) \cdot \alpha_{\mathcal{G}}(\mathbf{x})
    +\varphi c_{\mathcal{T}}(\mathbf{x}) \cdot \alpha_{\mathcal{T}}(\mathbf{x})}{\epsilon
    +\mu \alpha_{\mathcal{G}}(\mathbf{x})
    +\varphi \alpha_{\mathcal{T}}(\mathbf{x})}, \label{eq:color}
\end{align}
\end{subequations}
where we set $\mu = \varphi = 1$ and $\epsilon = 1e-6$.

For a fixed set of 2D viewing angles, NeRF's color output is determined by the 3D location, \ie, $F_{\Theta}(\mathbf{x} | \mathbf{d}) \rightarrow \mathbf{c}$. Drawing inspiration from CLIP-NeRF\cite{wang2022clip} and DreamFields\cite{jain2022zero}, we utilize Pose Sampling\cite{jain2022zero} to enhance NeRF's training, which leads to faster rendering speeds and improved quality. Pose Sampling involves randomly selecting camera poses for scene generation, with the ray's center being adjusted to the ROI $\Omega$ centroid to ensure better focus. By integrating \cref{eq:alpha}, \cref{eq:color}, and \cref{eq:nerf_1}, we produce 2D images $I_{\mathcal{S}}$, $I_{\mathcal{T}}$, and $I_{\mathcal{G}}$ from $\mathcal{S}_{\Theta}$, $\mathcal{T}_{\Theta}$, and $\mathcal{G}_{\Theta}$, all under the same viewing conditions. $I_{\mathcal{G}}$ is the final blend, with $I_{\mathcal{S}}$ and $I_{\mathcal{T}}$ serving as the source and target images for 2D blending.

\noindent\textbf{Configurations.} In our 5D Radiance Fields blending task, we adopt the architecture from the original NeRF \cite{mildenhall2021nerf} paper. We prepare source NeRF $\mathcal{S}_{\Theta}$ and target NeRF $\mathcal{T}_{\Theta}$, and create a clone $\mathcal{G}_{\Theta}$ from $\mathcal{S}_{\Theta}$ as the starting point for blending. During training, we use a Pose Sampling strategy, initializing a random camera pose $\mathbf{P}$ for each iteration and emitting $128 \times 128$ rays from this pose. Consequently, we obtain a $128 \times 128$ 2D image for each camera pose $\mathbf{P}$. We also employ the Adam optimizer \cite{kingma2017adam} with a cosine annealing strategy to gradually reduce the learning rate. The training is performed on a single NVIDIA RTX 3090 GPU, taking several hours to complete.

\noindent\textbf{Results.} \cref{fig:3d_result} showcases the blending results in various application contexts, comparing our method to the replacement blending strategy. This strategy directly substitutes the ROI $\Omega$ in $\mathcal{S}_{\Theta}$ with $c_{\mathcal{G}}(\mathbf{x})$, based on $\alpha_{\mathcal{G}}(\mathbf{x})$. The first two scenes illustrate a heterogeneous blending task with text effects, where text appears as if frozen in ice and submerged in water, achieved using the Neural Poisson Solver. The third scene depicts a large-scale blending task, featuring a chessboard encased in ice. 

Due to the limitations of graphics card capabilities and training duration, the resolution of our final rendered images in the 5D Radiance Fields blending task is capped at $256 \times 256$. This limitation may affect the visual quality of the blending results. Nonetheless, our method provides more coherent and natural blending results compared to the direct replacement strategy.

\begin{figure}[tb]
\centering
\includegraphics[width=\textwidth]{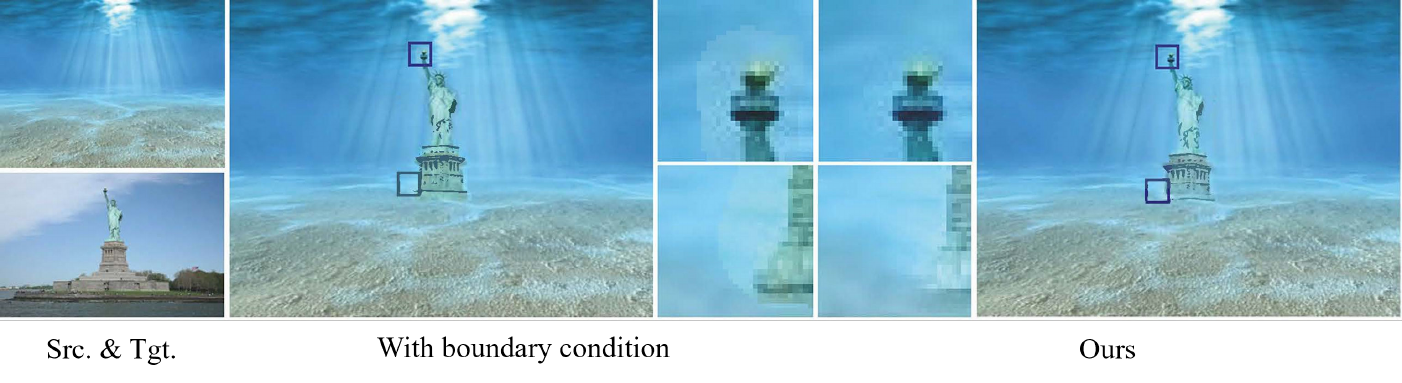}
\caption{\textbf{Ablation Experiment}. The control model using the boundary condition's local $\mathcal{L}_{\text{color}}$ is essentially similar to the method proposed by Pérez \etal \cite{perez2023poisson}, which performs scene blending only on $\partial \Omega$. This approach leads to background color distortion. Furthermore, since the $\mathcal{L}_{\text{color}}$ focuses on pixel values at $\partial \Omega$, it encounters difficulties in handling background textures. On the right, our method extends the pixel receptive field of $\mathcal{L}_{\text{color}}$ from the local $\partial \Omega$ to the global $(\mathcal{S} - \Omega)$, thereby enhancing background color recovery and achieving smooth transitions in the blending region.}
\label{fig:ablation}
\end{figure}

\subsection{Ablation Study}
\label{sec:ablation}
As outlined in \cref{sec:method_2d}, we discussed the research by Pérez \etal \cite{perez2023poisson}, in which they transformed the initial variational problem \cref{eq:poisson_1} into a Poisson equation \cref{eq:poisson_2}. This transformation was achieved through the integration of Dirichlet boundary conditions for a unique solution. In our pursuit to improve global blending effects and mitigate color blending errors often observed with traditional methods based on the Poisson equation, we extended this blending task to the entire background, represented as $(\mathcal{S} - \Omega)$. To evaluate the effectiveness of this approach, we conducted a series of comparative experiments to assess its impact on the final blending results.

To ensure experimental fairness in our control experiments, we followed the procedure of randomly initializing the INR, denoted as $\mathcal{G}$, which includes the pre-training blending results in the model using \cref{eq:loss_b}. In our method, we employed our proposed global $\mathcal{L}_{\text{color}}$ based on the entire background region. In the control model, we adjusted \cref{eq:loss_b} to depend solely on the local $\mathcal{L}_{\text{color}}$ at the blending boundary:
\begin{equation}
\label{eq:loss_b_change}
    \mathcal{L}_{\text{color}} 
    =\iint_{\partial \Omega}\|c_{\mathcal{G}}(\mathbf{x}) 
    -c_{\mathcal{S}}(\mathbf{x})\|^{2}.
\end{equation}

Our ablation study results are presented in \cref{fig:ablation}. From the results exhibited by the control model, it can be inferred that focusing only on $\partial \Omega$ leads to incorrect texture results at ignored locations. In contrast, our improved method effectively recovers the background and maintains natural transitions.

\subsection{Comparisons}
\label{sec:comparison}
\begin{figure}[tb]
\centering
\includegraphics[width=\textwidth]{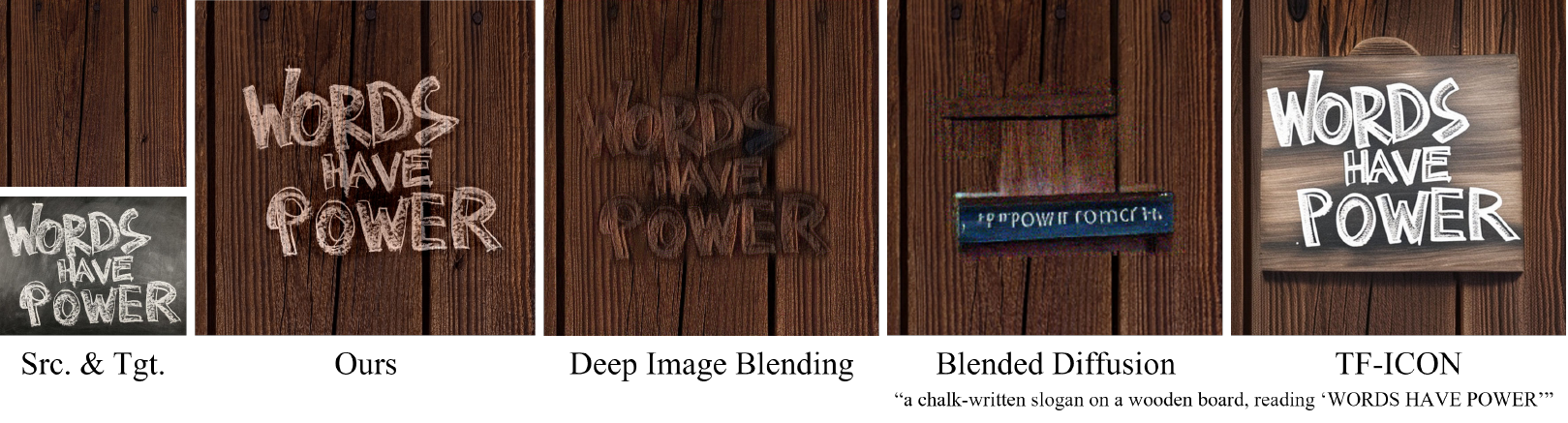}
\caption{Compared with generative models, our method offers greater controllability and allows signal blending with unrestricted dimensions.}
\label{fig:compare_with_gen}
\end{figure}

We compared our method with existing state-of-the-art approaches, as illustrated in \cref{fig:compare_with_gen}. \cite{zhang2020deep} optimizes blending using style and content loss from a deep network, while \cite{Avrahami_2022_CVPR} and \cite{lu2023tf} use generative models requiring additional inputs like text prompts. In contrast, our approach enhances blending controllability without introducing extraneous information, adhering to "natural blending" principles. Furthermore, our method does not require specific dimension optimization, making it versatile for signal blending across various dimensions.

\section{Conclusion}
In this paper, we present a universal INR-based framework specifically designed to handle blending tasks across various domains. The framework capitalizes on our Neural Poisson Solver to guide the blending and generation of INRs, enabling natural blending between source and target scenes. 

Our proposed methodology has undergone extensive testing across numerous 2D and 3D scenes and tasks, effectively demonstrating its adaptability in handling different dimensions. As our method fundamentally relies on the Neural Poisson Solver for model optimization without requiring additional prior knowledge, we believe that this framework can be readily adapted to a broad range of applications with minimal adjustments.

Nonetheless, we recognize that our framework has its limitations. One notable shortcoming is the occurrence of unnatural transitions within the blending area, especially when using masks with sharp edges like rectangles or other geometric shapes, which inadvertently makes the mask's presence more apparent. Another constraint surfaces in 5D Radiance Fields blending task where our Neural Poisson Solver requires three NeRF models of identical size for ray tracing throughout the blending process. This stipulation places substantial demands on GPU memory resources and subsequently affects training efficiency.

Moving forward, we aim to explore effective optimization strategies to enhance our architecture and expedite the model blending processes. We also plan to investigate blending tasks within more complex environments. We are optimistic about integrating these potential optimization methods with our proposed universal framework and look forward to their combined potential in future research pursuits.

%
%

\begin{thebibliography}{10}
\providecommand{\url}[1]{\texttt{#1}}
\providecommand{\urlprefix}{URL }
\providecommand{\doi}[1]{https://doi.org/#1}

\bibitem{Avrahami_2022_CVPR}
Avrahami, O., Lischinski, D., Fried, O.: Blended diffusion for text-driven editing of natural images. In: Proceedings of the IEEE/CVF Conference on Computer Vision and Pattern Recognition (CVPR). pp. 18208--18218 (June 2022)

\bibitem{benaim2024volumetric}
Benaim, S., Warburg, F., Christensen, P.E., Belongie, S.: Volumetric disentanglement for 3d scene manipulation. In: Proceedings of the IEEE/CVF Winter Conference on Applications of Computer Vision. pp. 8667--8677 (2024)

\bibitem{chen2022upst}
Chen, Y., Yuan, Q., Li, Z., Liu, Y., Wang, W., Xie, C., Wen, X., Yu, Q.: Upst-nerf: Universal photorealistic style transfer of neural radiance fields for 3d scene. arXiv preprint arXiv:2208.07059  (2022)

\bibitem{chen2020physics}
Chen, Y., Lu, L., Karniadakis, G.E., Dal~Negro, L.: Physics-informed neural networks for inverse problems in nano-optics and metamaterials. Optics express  \textbf{28}(8),  11618--11633 (2020)

\bibitem{chiang2022stylizing}
Chiang, P.Z., Tsai, M.S., Tseng, H.Y., Lai, W.S., Chiu, W.C.: Stylizing 3d scene via implicit representation and hypernetwork. In: Proceedings of the IEEE/CVF Winter Conference on Applications of Computer Vision. pp. 1475--1484 (2022)

\bibitem{dessein2014seamless}
Dessein, A., Smith, W.A., Wilson, R.C., Hancock, E.R.: Seamless texture stitching on a 3d mesh by poisson blending in patches. In: 2014 IEEE International Conference on Image Processing (ICIP). pp. 2031--2035. IEEE (2014)

\bibitem{elder1998image}
Elder, J.H., Goldberg, R.M.: Image editing in the contour domain. In: Proceedings. 1998 IEEE Computer Society Conference on Computer Vision and Pattern Recognition (Cat. No. 98CB36231). pp. 374--381. IEEE (1998)

\bibitem{fan2022unified}
Fan, Z., Jiang, Y., Wang, P., Gong, X., Xu, D., Wang, Z.: Unified implicit neural stylization. In: European Conference on Computer Vision. pp. 636--654. Springer (2022)

\bibitem{fattal2023gradient}
Fattal, R., Lischinski, D., Werman, M.: Gradient domain high dynamic range compression. In: Seminal Graphics Papers: Pushing the Boundaries, Volume 2, pp. 671--678 (2023)

\bibitem{gao2022objectfolder}
Gao, R., Si, Z., Chang, Y.Y., Clarke, S., Bohg, J., Fei-Fei, L., Yuan, W., Wu, J.: Objectfolder 2.0: A multisensory object dataset for sim2real transfer. In: Proceedings of the IEEE/CVF conference on computer vision and pattern recognition. pp. 10598--10608 (2022)

\bibitem{gong2023recolornerf}
Gong, B., Wang, Y., Han, X., Dou, Q.: Recolornerf: Layer decomposed radiance field for efficient color editing of 3d scenes. arXiv preprint arXiv:2301.07958  (2023)

\bibitem{gordon2023blended}
Gordon, O., Avrahami, O., Lischinski, D.: Blended-nerf: Zero-shot object generation and blending in existing neural radiance fields. arXiv preprint arXiv:2306.12760  (2023)

\bibitem{huang2022stylizednerf}
Huang, Y.H., He, Y., Yuan, Y.J., Lai, Y.K., Gao, L.: Stylizednerf: consistent 3d scene stylization as stylized nerf via 2d-3d mutual learning. In: Proceedings of the IEEE/CVF Conference on Computer Vision and Pattern Recognition. pp. 18342--18352 (2022)

\bibitem{jain2022zero}
Jain, A., Mildenhall, B., Barron, J.T., Abbeel, P., Poole, B.: Zero-shot text-guided object generation with dream fields. In: Proceedings of the IEEE/CVF Conference on Computer Vision and Pattern Recognition. pp. 867--876 (2022)

\bibitem{karniadakis2021physics}
Karniadakis, G.E., Kevrekidis, I.G., Lu, L., Perdikaris, P., Wang, S., Yang, L.: Physics-informed machine learning. Nature Reviews Physics  \textbf{3}(6),  422--440 (2021)

\bibitem{kingma2017adam}
Kingma, D.P., Ba, J.: Adam: A method for stochastic optimization (2017)

\bibitem{kobayashi2022decomposing}
Kobayashi, S., Matsumoto, E., Sitzmann, V.: Decomposing nerf for editing via feature field distillation. Advances in Neural Information Processing Systems  \textbf{35},  23311--23330 (2022)

\bibitem{lewis2001lifting}
Lewis, J.: Lifting detail from darkness. SIGGRAPH (2001)

\bibitem{liu2022recovery}
Liu, R., Sun, Y., Zhu, J., Tian, L., Kamilov, U.S.: Recovery of continuous 3d refractive index maps from discrete intensity-only measurements using neural fields. Nature Machine Intelligence  \textbf{4}(9),  781--791 (2022)

\bibitem{liu2021editing}
Liu, S., Zhang, X., Zhang, Z., Zhang, R., Zhu, J.Y., Russell, B.: Editing conditional radiance fields. In: Proceedings of the IEEE/CVF international conference on computer vision. pp. 5773--5783 (2021)

\bibitem{lu2023tf}
Lu, S., Liu, Y., Kong, A.W.K.: Tf-icon: Diffusion-based training-free cross-domain image composition. In: Proceedings of the IEEE/CVF International Conference on Computer Vision. pp. 2294--2305 (2023)

\bibitem{ma2023cardiacfield}
Ma, Z., Shen, C., Zhu, H., Zhou, Y., Liu, Y., Yi, S., Dong, L., Zhao, W., Brady, D., Cao, X., et~al.: Cardiacfield: Computational echocardiography for universal screening  (2023)

\bibitem{mildenhall2021nerf}
Mildenhall, B., Srinivasan, P.P., Tancik, M., Barron, J.T., Ramamoorthi, R., Ng, R.: Nerf: Representing scenes as neural radiance fields for view synthesis. Communications of the ACM  \textbf{65}(1),  99--106 (2021)

\bibitem{morel2012fourier}
Morel, J.M., Petro, A.B., Sbert, C.: Fourier implementation of poisson image editing. Pattern Recognition Letters  \textbf{33}(3),  342--348 (2012)

\bibitem{park2019deepsdf}
Park, J.J., Florence, P., Straub, J., Newcombe, R., Lovegrove, S.: Deepsdf: Learning continuous signed distance functions for shape representation. In: Proceedings of the IEEE/CVF conference on computer vision and pattern recognition. pp. 165--174 (2019)

\bibitem{perez2023poisson}
P{\'e}rez, P., Gangnet, M., Blake, A.: Poisson image editing. ACM Transactions on Graphics (TOG)  \textbf{22}(3),  313--318 (2003)

\bibitem{raissi2019physics}
Raissi, M., Perdikaris, P., Karniadakis, G.E.: Physics-informed neural networks: A deep learning framework for solving forward and inverse problems involving nonlinear partial differential equations. Journal of Computational physics  \textbf{378},  686--707 (2019)

\bibitem{raissi2020hidden}
Raissi, M., Yazdani, A., Karniadakis, G.E.: Hidden fluid mechanics: Learning velocity and pressure fields from flow visualizations. Science  \textbf{367}(6481),  1026--1030 (2020)

\bibitem{rocchini1999multiple}
Rocchini, C., Cignoni, P., Montani, C., Scopigno, R.: Multiple textures stitching and blending on 3d objects. In: Rendering Techniques’ 99: Proceedings of the Eurographics Workshop in Granada, Spain, June 21--23, 1999 10. pp. 119--130. Springer (1999)

\bibitem{ruckert2022neat}
R{\"u}ckert, D., Wang, Y., Li, R., Idoughi, R., Heidrich, W.: Neat: Neural adaptive tomography. ACM Transactions on Graphics (TOG)  \textbf{41}(4),  1--13 (2022)

\bibitem{shen2021non}
Shen, S., Wang, Z., Liu, P., Pan, Z., Li, R., Gao, T., Li, S., Yu, J.: Non-line-of-sight imaging via neural transient fields. IEEE Transactions on Pattern Analysis and Machine Intelligence  \textbf{43}(7),  2257--2268 (2021)

\bibitem{sitzmann2020implicit}
Sitzmann, V., Martel, J., Bergman, A., Lindell, D., Wetzstein, G.: Implicit neural representations with periodic activation functions. Advances in neural information processing systems  \textbf{33},  7462--7473 (2020)

\bibitem{sitzmann2021light}
Sitzmann, V., Rezchikov, S., Freeman, B., Tenenbaum, J., Durand, F.: Light field networks: Neural scene representations with single-evaluation rendering. Advances in Neural Information Processing Systems  \textbf{34},  19313--19325 (2021)

\bibitem{strumpler2022implicit}
Str{\"u}mpler, Y., Postels, J., Yang, R., Gool, L.V., Tombari, F.: Implicit neural representations for image compression. In: European Conference on Computer Vision. pp. 74--91. Springer (2022)

\bibitem{tancik2020fourier}
Tancik, M., Srinivasan, P., Mildenhall, B., Fridovich-Keil, S., Raghavan, N., Singhal, U., Ramamoorthi, R., Barron, J., Ng, R.: Fourier features let networks learn high frequency functions in low dimensional domains. Advances in Neural Information Processing Systems  \textbf{33},  7537--7547 (2020)

\bibitem{tewari2022advances}
Tewari, A., Thies, J., Mildenhall, B., Srinivasan, P., Tretschk, E., Yifan, W., Lassner, C., Sitzmann, V., Martin-Brualla, R., Lombardi, S., et~al.: Advances in neural rendering. In: Computer Graphics Forum. vol.~41, pp. 703--735. Wiley Online Library (2022)

\bibitem{wang2022clip}
Wang, C., Chai, M., He, M., Chen, D., Liao, J.: Clip-nerf: Text-and-image driven manipulation of neural radiance fields. In: Proceedings of the IEEE/CVF Conference on Computer Vision and Pattern Recognition. pp. 3835--3844 (2022)

\bibitem{wu2022palettenerf}
Wu, Q., Tan, J., Xu, K.: Palettenerf: Palette-based color editing for nerfs. arXiv preprint arXiv:2212.12871  (2022)

\bibitem{xu2023nesvor}
Xu, J., Moyer, D., Gagoski, B., Iglesias, J.E., Grant, P.E., Golland, P., Adalsteinsson, E.: Nesvor: Implicit neural representation for slice-to-volume reconstruction in mri. IEEE Transactions on Medical Imaging  (2023)

\bibitem{yuan2022nerf}
Yuan, Y.J., Sun, Y.T., Lai, Y.K., Ma, Y., Jia, R., Gao, L.: Nerf-editing: geometry editing of neural radiance fields. In: Proceedings of the IEEE/CVF Conference on Computer Vision and Pattern Recognition. pp. 18353--18364 (2022)

\bibitem{zhang2020deep}
Zhang, L., Wen, T., Shi, J.: Deep image blending. In: The IEEE Winter Conference on Applications of Computer Vision. pp. 231--240 (2020)

\bibitem{zhou2023fourier}
Zhou, H., Feng, B.Y., Guo, H., Lin, S.S., Liang, M., Metzler, C.A., Yang, C.: Fourier ptychographic microscopy image stack reconstruction using implicit neural representations. Optica  \textbf{10}(12),  1679--1687 (2023)

\bibitem{zhu2022dnf}
Zhu, H., Liu, Z., Zhou, Y., Ma, Z., Cao, X.: Dnf: diffractive neural field for lensless microscopic imaging. Optics Express  \textbf{30}(11),  18168--18178 (2022)

\bibitem{zhu2023disorder}
Zhu, H., Xie, S., Liu, Z., Liu, F., Zhang, Q., Zhou, Y., Lin, Y., Ma, Z., Cao, X.: Disorder-invariant implicit neural representation. arXiv preprint arXiv:2304.00837  (2023)

\bibitem{zhu2023pyramid}
Zhu, J., Zhu, H., Zhang, Q., Zhu, F., Ma, Z., Cao, X.: Pyramid nerf: Frequency guided fast radiance field optimization. International Journal of Computer Vision pp. 1--16 (2023)

\end{thebibliography}

\end{document}